\documentclass[letterpaper, 10 pt, conference]{ieeeconf}  
\IEEEoverridecommandlockouts                              
\usepackage{graphicx}
\usepackage{epsfig} 
\usepackage{times} 
\usepackage{amsmath} 
\usepackage{amssymb}  
\usepackage{subfigure}

\newtheorem{theorem}{Theorem}
\newtheorem{lemma}{Lemma}

\usepackage{algorithm}
\usepackage{algorithmicx}
\usepackage{algpseudocode}

\usepackage[american]{babel}

\title{\LARGE \bf
Efficient Globally-Optimal Correspondence-Less\\Visual Odometry for Planar Ground Vehicles
}

\author{Ling Gao, Junyan Su, Jiadi Cui, Xiangchen Zeng, Xin Peng, and Laurent Kneip
\thanks{All the authors are with the School of School of Information Science and Technology, ShanghaiTech University. 
{\tt\small \{gaoling, sujy, cuijd, zengxch, pengxin1, lkneip\}@shanghaitech.edu.cn}.}
}

\begin{document}

\maketitle
\thispagestyle{empty}
\pagestyle{empty}


\begin{abstract}
The motion of planar ground vehicles is often non-holonomic, and as a result may be modelled by the 2 DoF Ackermann steering model. We analyse the feasibility of estimating such motion with a downward facing camera that exerts fronto-parallel motion with respect to the ground plane. This turns the motion estimation into a simple image registration problem in which we only have to identify a 2-parameter planar homography. However, one difficulty that arises from this setup is that ground-plane features are indistinctive and thus hard to match between successive views. We encountered this difficulty by introducing the first globally-optimal, correspondence-less solution to plane-based Ackermann motion estimation. The solution relies on the branch-and-bound optimisation technique. Through the low-dimensional parametrisation, a derivation of tight bounds, and an efficient implementation, we demonstrate how this technique is eventually amenable to accurate real-time motion estimation. We prove its property of global optimality and analyse the impact of assuming a locally constant centre of rotation. Our results on real data finally demonstrate a significant advantage over the more traditional, correspondence-based hypothesise-and-test schemes.
\end{abstract}


\section{INTRODUCTION}

Accurate velocity measurements are an essential ingredient to the stable localisation and control of ground vehicles. A simple forward integration of vehicle speed can provide an estimate of the vehicle trajectory, a procedure commonly known as \textit{dead-reckoning}. Although the forward integration will lead to a random walk, fusing its value in a complementary filter with an absolute reference signal such as GPS may already be sufficient to attain reasonably accurate localisation. However, while the most straightforward solution to vehicle speed estimation is given by employing wheel odometers, the risk of varying and unknown wheel parameters as well as hard-to-model wheel slippage have ever since motivated the use of contact-less \textit{visual odometry} to measure incremental vehicle displacements \cite{moravec80}. Our paper focuses on this problem.

Our work is inspired by optical mouse sensors. The idea is that for any device that is being moved over a flat, planar surface, the velocity in the plane can simply be measured by an optical sensor that directly faces the plane. As the device is moving, the displacement information is deduced from the apparent motion of brightness patterns in the perceived image. The important motivation for a downward facing sensor is that the depth and structure of the scene will be known in advance, and that---as a result---the motion of patterns in the image under planar displacement can be described by a simple Euclidean transformation. However, even though this estimation problem appears to be simple, it is difficult to find point correspondences between subsequent images as the perceived floor texture often does not lead to distinctive, easily matchable keypoints.

Our work aims at providing an answer to this problem, notably by the following three contributions:
\begin{itemize}
\item We propose to solve the above described registration problem using a correspondence-free approach. Given point sets measured in two images,  the algorithm searches the space of all possible planar displacements such that the number of correspondences is maximised. The definition of a correspondence does not depend on descriptor information, but is merely based on whether or not the Euclidean distance between a candidate pair of points drops below a certain threshold.
\item We present an efficient globally optimal solution to this problem by employing the branch-and-bound optimisation technique. We present tight bounds derived from our original optimisation objective, and an implementation that is eventually amenable to real-time execution.
\item The complexity of the branch-and-bound search strategy is further reduced by changing to a two degree-of-freedom parametrisation of the non-holonomic displacement of planar ground vehicles, the \textit{Ackermann} model.
\end{itemize}
To the best of our knowledge, we are the first to introduce a globally optimal image registration technique for fronto-parallel motion in front of a planar scene and obeying the \textit{Ackermann steering} model. Our experiments suggest that this strategy is able to solve challenging cases for which image-to-image correspondences are hard to derive, and we further demonstrate that the low-degree parametrisation of this problem still enables near real-time image processing, a feature that is rarely achieved for branch-and-bound optimisation methods. Our paper is structured as follows. Firstly, we introduce further related work in Section \ref{sec:related_work}. Theoretical foundations such as planar homography and the Ackermann steering model are briefly reviewed in Section \ref{sec:foundations}. Our contribution of a correspondence-less, globally optimal registration method is introduced in Section \ref{sec:branch_and_bound}. Sections \ref{sec:algorithm} and \ref{sec:experiments} finally conclude with outlining our implementation and successful application to both simulated and real data.


\section{RELATED WORK}
\label{sec:related_work}

The image displacements for a camera moving in front of a planar scene can be modelled by a simple homography, which furthermore has a closed algebraic form as a function of the relative displacement, the camera intrinsics, and the plane parameters \cite{hartley2003multiple,ma2012invitation}. A homography is an 8 DoF image-to-image transformation, and therefore generally requires 4 points to be solved. The complexity of the solution may be reduced by assumption of known plane parameters, or even constraints on the motion such as for example planar motion. For a downward motion-plane facing camera, these conditions would reduce the problem to a simple 3 DoF Euclidean transformation estimation. Other related work is given by \cite{scaramuzza2009real} and \cite{huang2019motion}, which substitute the Ackermann steering model into the essential matrix or an $n$-linearity. The imposition of epipolar incidence relationships that do no longer require either structure or scale parameters to be derived makes it possible for those methods to identify the relative displacement based on a single feature correspondence.

The aforementioned solutions are interesting and related in that they exploit structure or motion related priors to solve the registration problem. However, they require feature correspondences for which at least a majority is consistent with a single dominant camera displacement. The dominant motion is then identified by sampling and testing hypotheses within a heuristic framework \cite{fischler81}, which is why such methods cannot guarantee to find the optimal inlier set and an associated image-to-image transformation. Ransac-based methods contrast with globally optimal solutions that search the entire space of possible transformations to identify the optimal inlier set, possibly even without the prior requisite of point correspondences. The most common method here that has been regularly applied in geometric computer vision is given by branch-and-bound optimisation \cite{land2010automatic}.

Branch-and-bound has been used to solve correspondence-less 2D-3D \cite{brown2015globally,brown2019family,liu20182d,campbell2017globally,campbell2018globally,campbell2019alignment}, 2D-2D \cite{hartley2007global,hartley2009global,li2009consensus,neveu2019generic}, and 3D-3D \cite{li20073d,yang2013go,parra2014fast,campbell2016gogma,liu2018efficient} registration problems. A very early solution to 2D-3D registration similar to branch-and-bound has been introduced in \cite{jurie1999solution}, while \cite{brown2019family} provides a decent summary on globally optimal, correspondence-less camera resectioning. It is common for earlier works to add prior knowledge or restrict the optimisation domain to improve computational efficiency. \cite{olsson2008branch} requires known correspondences, while \cite{brown2015globally} requires prior knowledge on the inlier fraction to trim outliers. \cite{li20073d,hartley2007global,parra2014fast,liu20182d} reduce the optimisation to a search in the space of rotations, while \cite{liu2018efficient} exploits rotation invariant features to reduce the search to translations. Branch-and-bound has also been used to solve other, related problems, such as  medical image registration \cite{pfeuffer2012discrete}, auto-calibration \cite{paudel2018sampling}, or motion estimation for multi-camera systems \cite{kim2008motion}.

Another line of research that is related to ours is globally optimal image matching for sparse \cite{mount1999efficient,breuel2003implementation,bazin2012branch} or even semantic, region-based \cite{speciale2018consensus} features. Such methods proceed by branching in the space of a low-dimensional image-to-image mapping, for example a 4 DoF transformation in the case of \cite{breuel2003implementation}. An alternative for Euclidean image registration that does not require the extraction of sparse features is given by the application of the Fourier Mellin transform \cite{guo05,bulow2009}, which provides high computational efficiency but no guarantees of global optimality.

To the best of our knowledge, we present the first correspondence-less globally optimal image-to-image registration method which is rooted in the non-holonomic motion model of ground plane vehicles, thus leading to a formulation in only 2 DoF and near real-time performance.


\section{FOUNDATIONS}
\label{sec:foundations}

We briefly review the geometry of the Ackermann steering model and its embedding into a planar homography for a vehicle-mounted, downward-facing camera. The section concludes with a presentation of the optimisation objective in a correspondence-less scenario.

\subsection{The Ackermann steering model}

A connection between the relative rotation and translation of a ground vehicle obeying the Ackermann steering model has been originally found in \cite{scaramuzza2009real}. It is a common simplification to reduce the dimensionality of the problem from 3 DoF for general planar motion to only 2 DoF.
\begin{figure}[b]
	\center
	\includegraphics[width = 0.49\textwidth]{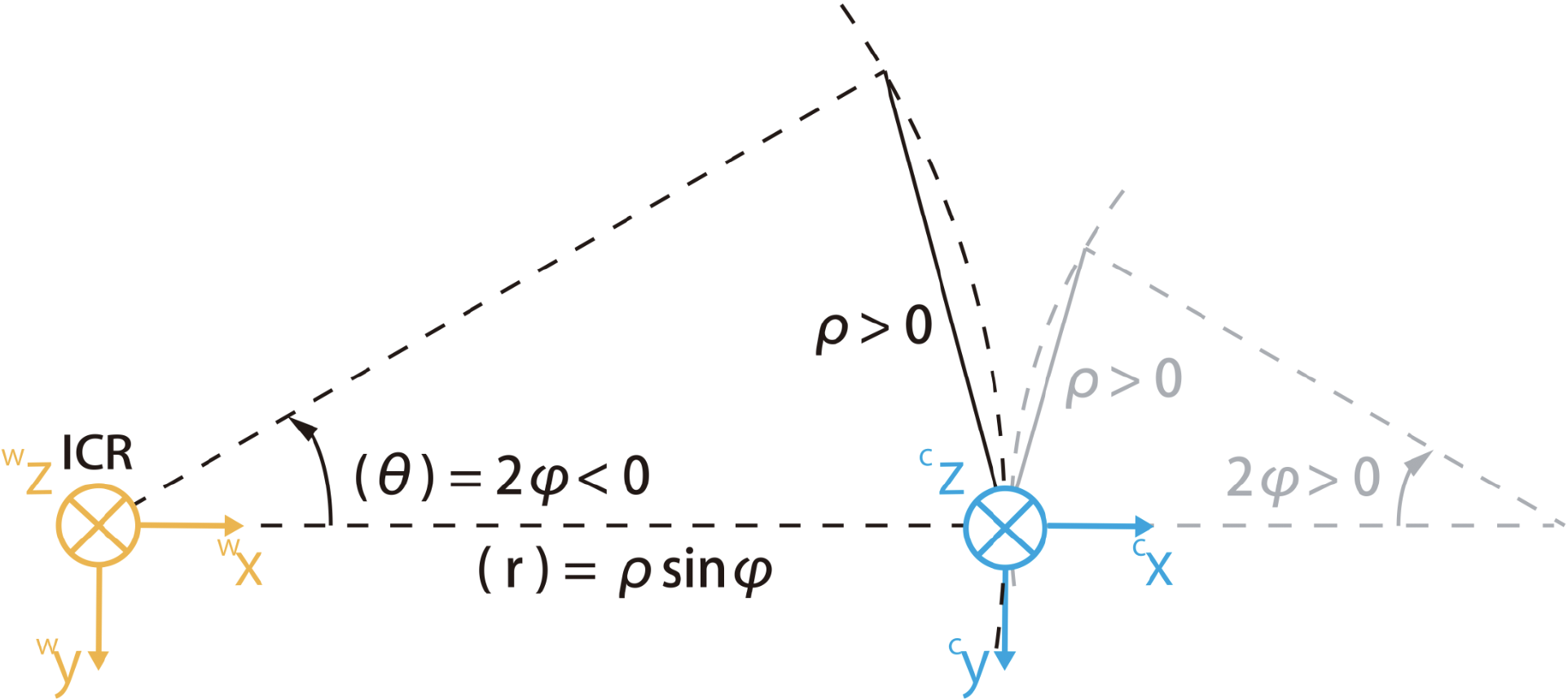} 
	\caption{\textit{The Ackermann steering model:} Two scenarios are illustrated here: left turn and right turn, for which $\phi$ takes on a negative or positive value.}
	\label{Ackermann Steering Model}
\end{figure}
As illustrated in Figure \ref{Ackermann Steering Model}, the movement of a ground vehicle is approximated by a circular arc about an instantaneous centre of rotation, which degenerates into a line if the vehicle simply moves forward and the radius becomes infinite. Whenever the vehicle is obeying this model, its heading remains tangential to the arc, which explains the non-holonomic coupling between rotation and translation. The instantaneous centre of rotation is the point in which all wheels' instantaneous rolling axes are intersecting.

A natural choice for 2 DoF would be rotation angle $\theta$ and arc-radius $r$. However, this would lead to problems in a branch-and-bound setting, as the algorithm would have to branch over the radius $r$ that could potentially take on infinite values if heading straight. Therefore, as originally suggested in \cite{scaramuzza2009real}, we will use the half angle $\phi$ and the baseline $\rho$ to describe the motion between successive frames. The relationship between the two sets of parameters can be expressed by $\theta = 2 \phi$ and $r = \rho \sin(\phi)$. We will adopt the convention that the vehicle frame coincides with the frame of the downward facing camera. The $y$ axis is pointing in the backwards direction, the $z$ axis downward, and the $x$ axis to the right. Under this convention, the relative rotation matrix $\mathbf{R}$ and translation vector $\mathbf{t}$ can be denoted as
\begin{equation}
	\label{Rotation_Matrix}
	\textbf{R} = \left[
    				\begin{matrix}
						\cos\!\left(2\phi\right) & - \sin\!\left(2\phi\right) & 0 \\
						\sin\!\left(2\phi\right) &   \cos\!\left(2\phi\right) & 0 \\
						0 & 0 & 1
	  				\end{matrix}
  			   	 \right] \text{ and }
	\textbf{t} = \rho \left[
    					\begin{matrix}
			   				  \sin\!\left(\phi\right) \\
			   				- \cos\!\left(\phi\right) \\
			   				                        0
  						\end{matrix}
  					  \right].
\end{equation}
It is easy to see that positive, negative, or zero values for $\phi$ denote a right turn, a left turn, or forward motion, respectively. $\rho$ stays a positive value as long as the vehicle moves in the forward direction.

\subsection{The planar homography model}

As the planar motion model introduced in the previous section suggests, we may avoid further complexity caused by unknown depth-of-scene. The camera faces downwards as in an aerial imaging scenario, and therefore observes the motion plane for which the normal vector points in the opposite direction of the camera principal axis \cite{hartley2003multiple,ma2012invitation}. We further assume that the motion is perfectly planar and that the depth of the plane is perfectly known. The planar homography therefore becomes the image-to-image transformation matrix,
\begin{equation}
    \label{homography}
    \textbf{H} = \textbf{K}(\textbf{R}-\frac{\textbf{t}\textbf{n}^\textbf{T}}{d})\textbf{K}^{\textbf{-1}},
\end{equation}
which---if using the same relative rotation parameters $\mathbf{R}$ and $\mathbf{t}$ as introduced in the previous section---permits the transformation of points from the second back to the first frame. $\mathbf{K}$ denotes the intrinsic camera matrix
\begin{equation}
    \label{camera_intrinsic_matrix}
    \textbf{K} = \left[
			        \begin{matrix}
			            a & 0 & u_0 \\
			            0 & a & v_0 \\
			            0 & 0 & 1
			        \end{matrix}
        		 \right],
\end{equation}
$\mathbf{n}=[0 \text{ } 0 \text{ } -1]^{T}$ is the unit normal of the plane, and $d$ is the depth of the plane.

Normally, at least 4 correspondence pairs are required to calculate the homography matrix. However, by substituting (\ref{Rotation_Matrix}) in (\ref{homography}), and considering the camera matrix $\textbf{K}$, the distance $d$, and the unit normal vector $\textbf{n}$ as known, we easily see that the homography is a function of only 2 unknown parameters, which are the baseline $\rho$ and the half-angle $\phi$. It may therefore be solved at the hand of only a single point correspondence between the two views. The homography furthermore takes on the form of the Euclidean transformation matrix
\begin{equation}
    \label{homography_matrix}
    \textbf{H} = \left[
			        \begin{matrix} 
			            h_1 & h_2 & h_3 \\
			            h_4 & h_5 & h_6 \\ 
			            0 & 0 & 1
			        \end{matrix}
    \right],
\end{equation}
with
\begin{equation}
    \label{homography_matrix_elements}
    \begin{aligned}
        h_1 &=   \cos\left(2\phi\right) \\
        h_2 &= - \sin\left(2\phi\right)\\
        h_3 &=   u_0 - u_0 \cos\left(2\phi\right) + v_0 \sin\left(2\phi\right) + \frac{a}{d}\ \rho\sin\left(\phi\right) \\
        h_4 &=   \sin\left(2\phi\right) \\
        h_5 &=   \cos\left(2\phi\right) \\
        h_6 &=   v_0 - v_0 \cos\left(2\phi\right) - u_0 \sin\left(2\phi\right) - \frac{a}{d}\ \rho\cos\left(\phi\right).
    \end{aligned}
\end{equation}

Points from the second view may hence be transformed to the original frame by using the simple relations
\begin{equation}
	\label{H_X}
	x_1 = [x_2 - u_0] cos(2 \phi) + [v_0 - y_2] sin(2 \phi) + \frac{a}{d} \rho sin(\phi) + u_0
\end{equation}
\begin{equation}
	\label{H_Y}
	y_1 = [x_2 - u_0] sin(2 \phi) + [y_2 - v_0] cos(2 \phi) - \frac{a}{d} \rho cos(\phi) + v_0.
\end{equation}

\subsection{Problem Formulation}

Our correspondence-less optimisation objective is given as
\begin{equation}
	\label{optimization}
	\sigma^* = \max\limits_{\textbf{H}} f(\textbf{H}) = \max\limits_{\textbf{R}, \textbf{t}} f(\textbf{R}, \textbf{t}) = \max\limits_{\phi, \rho} f(\phi, \rho) \text{, where} \nonumber
\end{equation}
\begin{equation}
	\label{objective_function}
	f(\textbf{H}) = \sum_{i \in \mathcal{P}_1} \sum_{j \in \mathcal{P}_2} 1\{\epsilon - ||p_{i}-\textbf{H} p_{j}|| \}.
\end{equation}
$\mathbf{p}_{i}$ and $\mathbf{p}_{j}$ refer to the homogeneous form of the local invariant keypoints in the first and second view, respectively. The objective consists of searching the space of admissible $\phi$ and $\rho$ to find optimal values for which the count of point pairs for which the Euclidean distance after the registration falls below a pre-defined threshold $\epsilon$ is maximised. $\epsilon$ acts as an acceptable inlier threshold to accommodate for noise in the feature point locations. Note that $1(\cdot)$ is a step function that returns one if the argument is bigger than zero, and zero otherwise. The fact that correspondences are counted purely based on geometric distance means that no time has to be invested into feature matching.


\section{BRANCH AND BOUND}
\label{sec:branch_and_bound}

Branch-and-bound optimization offers a solid globally-optimal perspective to solve a highly non-convex and possibly discrete maximisation problem \cite{land2010automatic}. The method consists of repeatedly branching over the domain of the optimisation parameters, each time finding lower and upper bounds on the maximisation objective. If the upper bound on the objective value in a certain interval remains below the lower bound in another interval, the first interval can be completely discarded as it impossibly contains the global maximum of the objective function. In each iteration, the algorithm effectively discards all such intervals, and branches the remaining intervals into smaller ones. The bounds in smaller intervals are generally tighter, which leads to the possibility of further intervals to be removed in each subsequent iteration. The algorithm converges to the globally optimal solution if a required precision is reached, or the upper and lower bounds become the same. In the following, we will see the derivation of the bounds.

\subsection{Lower Bound}

Any value of the objective function sampled within a certain interval of the optimisation parameters can be treated as a lower bound. This is clear as the global maximum within a certain interval must be either equal or better than the value retrieved from an arbitrary sample within the interval. In our implementation, we simply sample a value at the centre of each interval to retrieve a lower bound. Owing to the even branching of the optimisation space, the centre of an interval will always turn out to be on the boundary of the sub-intervals, which generates proper bisectioning and avoids redundant calculations.

\subsection{Upper Bound}

The efficiency of the branch and bound algorithm highly depends on both how much time it takes to calculate a single bound (i.e. the \textit{bound simplicity}) and how many intervals will have to be analysed at each level, which heavily depends on the \textit{bound tightness}. These are competing objectives. In order to achieve an upper bound that is both simple and tight, we distinguish between different scenarios of the vehicle motion: (A) heading straight, and (B) taking a turn. The two scenarios have 1 and 2 DoF, respectively.

(A) When the vehicle is heading straight, only 1 DoF from the translation contributes to the feature displacements. Equations (\ref{H_X}) and (\ref{H_Y}) can notably be simplified as:
\begin{equation}
	\label{H_simplified_X}
	x_1 = x_2
\end{equation}
\begin{equation}
	\label{H_simplified_Y}
	y_1 = y_2  + \frac{a}{d} \rho.
\end{equation}
This creates a tight bounding box for the transferred location of point $[x_2 \text{ } y_2]^T \in \mathcal{P}_2$, given by 
\begin{theorem}\ 
\textit{(Bounding box for the transfer location in the case of forward motion)}
\begin{equation}
	\label{UB_Head_Straight_X}
	\underline{x_1} = \overline{x_1} = x_2
\end{equation}
\begin{equation}
	\label{UB_Head_Straight_Y}
	\underline{y_1} = y_2  - \frac{a}{d} P_{max},\  \overline{y_1} = y_2  - \frac{a}{d} P_{min}
\end{equation}
Any point from the set $\mathcal{P}_1$ that lies within the interval $[\underline{x_1}-\epsilon,\overline{x_1}+\epsilon] \times [\underline{y_1}-\epsilon,\overline{y_1}+\epsilon]$ will contribute to the upper bound on the interval $[P_{min},P_{max}]$.
\end{theorem}

(B) When the vehicle is taking a turn, the situation becomes slightly more involved. To start with, we introduce useful inequalities based on the piece-wise monotonicity of trigonometric functions.

\begin{lemma} \ 
\textit{(Monotonicity of trigonometric functions)}

For an arbitrary degree $\phi \in [\Phi_1, \Phi_2] \subseteq [-\frac{\pi}{2}, \frac{\pi}{2}]$, 
\begin{equation}
	\sin(\Phi_1) \le \sin(\phi) \le \sin(\Phi_2).
\end{equation}

For an arbitrary degree $\phi \in [\Phi_3, \Phi_4] \subseteq [-\frac{\pi}{2}, 0]$, 
\begin{equation}
	\cos(\Phi_3) \le \cos(\phi) \le \cos(\Phi_4).
\end{equation}

For an arbitrary degree $\phi \in [\Phi_5, \Phi_6] \subseteq [0, \frac{\pi}{2}]$, 
\begin{equation}
	\cos(\Phi_6) \le \cos(\phi) \le \cos(\Phi_5).
\end{equation}
\end{lemma}

It is hard to put a simple, unique upper bound on Equations (\ref{H_X}) and (\ref{H_Y}) as $\cos(\phi)$ is non-monotonous over the entire interval $[-\frac{\pi}{2}, \frac{\pi}{2}]$, and also $(x_2-u_0)$ and $(v_0-y_2)$ may take on different signs depending on which quadrant with respect to the principal point the image point $[x_2 \text{ } y_2]^T$ lies in. Our approach consists of again distinguishing different scenarios. First, we distinguish between right and left turns for which $\phi>0$ and $\phi<0$, respectively. This is always possible if simply making sure that each examined interval $[\Phi_{min},\Phi_{max}]$ does not include 0 unless it is the boundary of the interval. Second, we consider the point $[x_2 \text{ } y_2]^T$ and which of the four quadrants it lies in. Taking all possible combinations, we obtain 8 different ways to compute a bounding box for the transfer location of a point $[x_2 \text{ } y_2]^T$ given a certain interval of possible motion parameters $[P_{min},P_{max}]\times[\Phi_{min},\Phi_{max}]$ (for matters of simplicity, we omit the case of backwards motion for which $\rho < 0$).

\begin{theorem}\
\textit{(Transfer bounding box for a right turn and a feature point in the lower left corner of the image)}

We have $x_2 < u_0$, $y_2 > v_0$, $\phi\in[\Phi_{min}, \Phi_{max}]$, and $\rho\in[P_{min}, P_{max}]$, with $\Phi_{min} > 0$ and $P_{min} > 0$. We obtain

\begin{equation}
	\label{UB_Turn_Right_X_min}
	\begin{aligned}
		\underline{x_1} &= [x_2 - u_0] cos(2 \Phi_{min}) + [v_0 - y_2] sin(2 \Phi_{max}) \\
						&\ \ \ + \frac{a}{d} P_{min} sin(\Phi_{min}) + u_0
	\end{aligned}				
\end{equation}
\begin{equation}
	\label{UB_Turn_Right_X_max}
	\begin{aligned}
		\overline{x_1} &= [x_2 - u_0] cos(2 \Phi_{max}) + [v_0 - y_2] sin(2 \Phi_{min}) \\
					   &\ \ \ + \frac{a}{d} P_{max} sin(\Phi_{max}) + u_0
	\end{aligned}
\end{equation}
\begin{equation}
	\label{UB_Turn_Right_Y_min}
	\begin{aligned}
		\underline{y_1} &= [x_2 - u_0] sin(2 \Phi_{max}) + [y_2 - v_0] cos(2 \Phi_{max}) \\
		   				&\ \ \ \ - \frac{a}{d} P_{max} cos(\Phi_{min}) + v_0
	\end{aligned}
\end{equation}
\begin{equation}
	\label{UB_Turn_Right_Y_max}
	\begin{aligned}
		\overline{y_1} &= [x_2 - u_0] sin(2 \Phi_{min}) + [y_2 - v_0] cos(2 \Phi_{min}) \\
	   				   &\ \ \ \ - \frac{a}{d} P_{min} cos(\Phi_{max}) + v_0.
	\end{aligned}
\end{equation}
Again, any point from the set $\mathcal{P}_1$ that lies within the interval $[\underline{x_1}-\epsilon,\overline{x_1}+\epsilon] \times [\underline{y_1}-\epsilon,\overline{y_1}+\epsilon]$ will contribute to the upper bound on the interval $[\Phi_{min}, \Phi_{max}]\times[P_{min},P_{max}]$. Note that case (B) includes case (A) as a special case, so prior knowledge about whether motion is straight is not required.

\end{theorem}

\begin{figure*}[t]
  \vspace{0.2cm}
  \centering
  \subfigure[Before non-linear refinement]{
    \centering
    \includegraphics[width=0.3\linewidth]{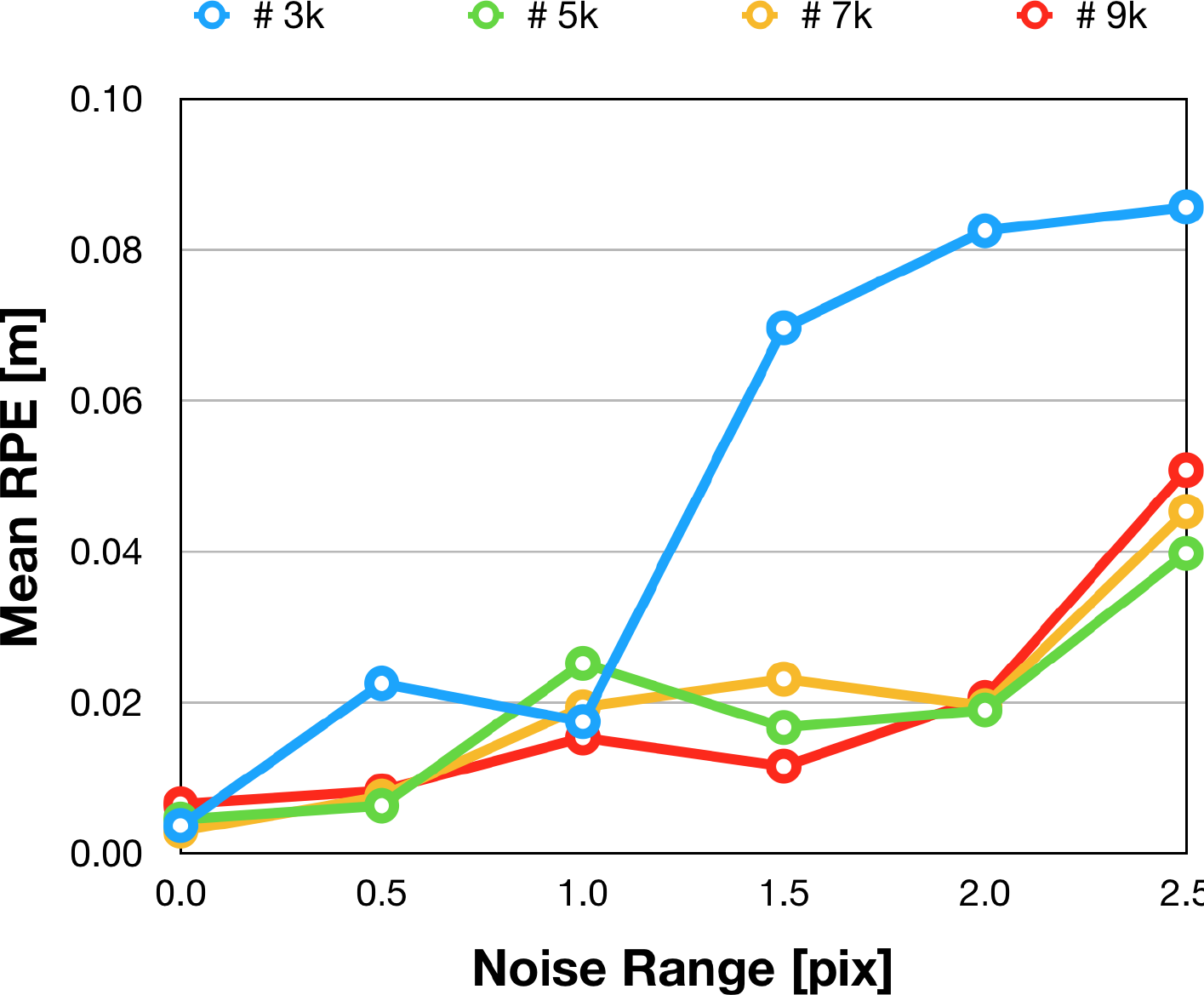}
    \label{fig:simulation:before}
  }
  \subfigure[After non-linear refinement]{
    \centering
    \includegraphics[width=0.3\linewidth]{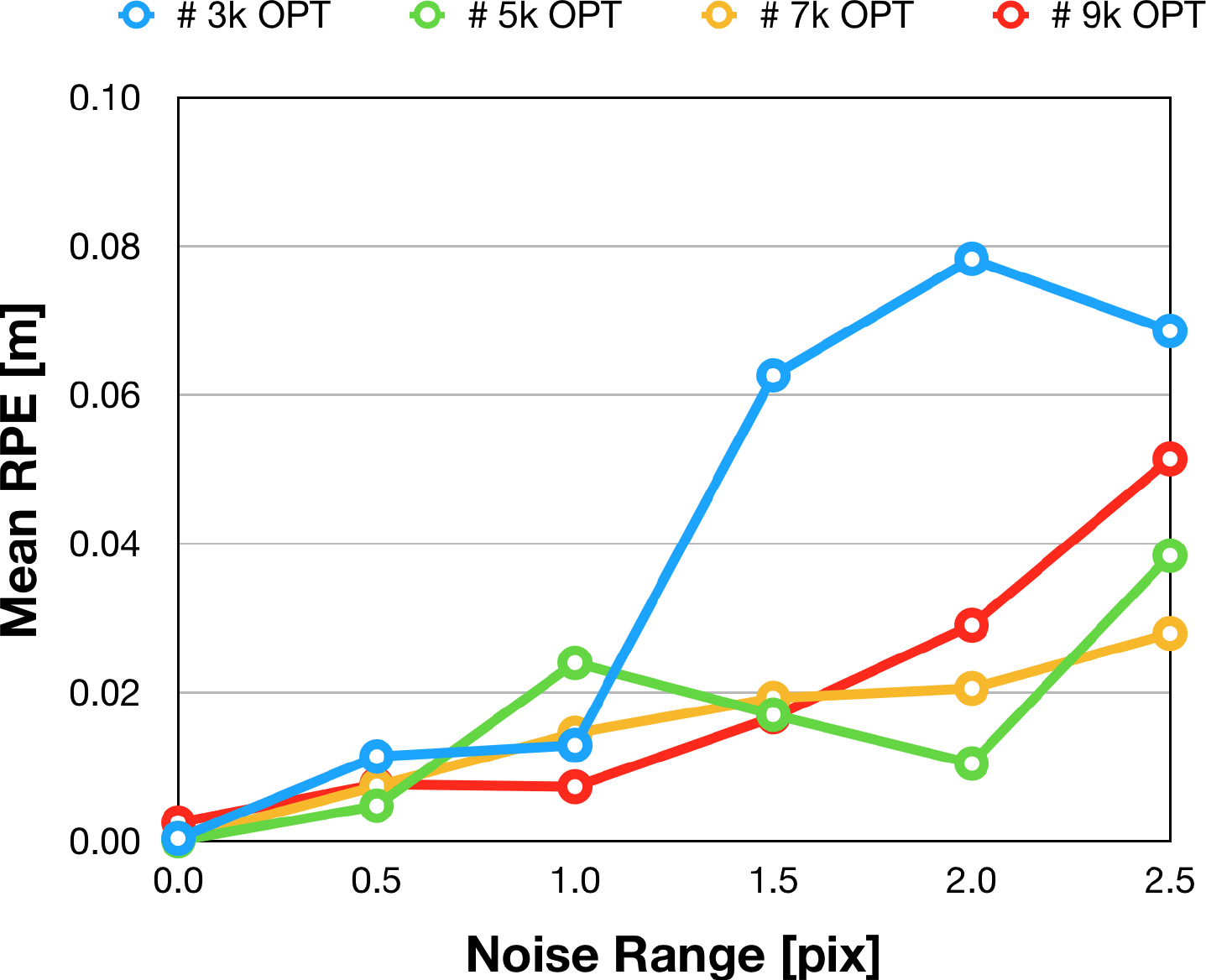}
    \label{fig:simulation:after}
  }
  \subfigure[With varying eccentricity]{
    \centering
    \includegraphics[width=0.3\linewidth]{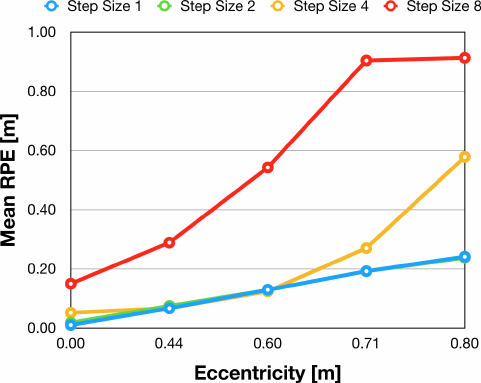}
    \label{fig:simulation:eccentricity}
  }
  \caption{Performance of our algorithm over synthetically generated correspondences with varying feature density and noise properties (left and centre) and displacements along non-circular arcs (right). The left and center figure show the result before and after non-linear refinement, respectively.}
  \label{fig:simulation}
\end{figure*}

\section{ALGORITHM IMPLEMENTATION}
\label{sec:algorithm}

Our overall algorithm denoted Globally-Optimal correspondence-less Visual Odometry (GOVO) is outlined in Algorithm \ref{GOVO_algorithm} and aided by the helper function Algorithm \ref{Space_Algorithm} to calculate bound values within a certain interval.

\begin{algorithm}
	\caption{GOVO: efficient branch and bound algorithm for globally-optimal correspondence-less visual odometry}
	\begin{algorithmic}[1]
		\Require point sets $\mathcal{P}_1$ and $\mathcal{P}_2$, initial domains $\Phi$ and P, camera intrinsic matrix $\textbf{K}$, distance between camera and ground $d$, acceptable inlier threshold $\epsilon$
		\Ensure optimal solution $(\phi^{\star}, \rho^{\star})$, 2D-2D correspondences
		
		\State Initialize a space $S_0$ over the domains $\Phi$ and P
		\State Execute Algorithm 2 for $S_0$
		\State Push $S_0$ into queue $Q$
		\Loop
			\State Pop a space $S'$ from $Q$
			\If {$S' \rightarrow  c^{\star} = 1$} 
				\State \textbf{for} space $S''$ at same level with $S'$ in $Q$ \textbf{then}
					\State \ \ \ \ \textbf{if} {$S'' \rightarrow c^{\star} = 1$} \textbf{then} Collect correspondences
				\State \textbf{end for}
				\State Local refinement over all inlier correspondences
				\State \textbf{return} refined result $(\phi^{\star}, \rho^{\star})$
			\Else
				\State Split $S$ into subspaces ${S_{1}, S_{2}, S_{3}, S_{4}}$
				\State Execute Algorithm 2 for ${S_{1}, S_{2}, S_{3}, S_{4}}$
				\State Prune and push into $Q$
			\EndIf
		\EndLoop
	\end{algorithmic}
	\label{GOVO_algorithm}
\end{algorithm}

\begin{algorithm}
	\caption{bound calculation}
	\begin{algorithmic}[1]
		\Require point sets $\mathcal{P}_1$ and $\mathcal{P}_2$, domains $\Phi$ and P, camera intrinsic matrix $\textbf{K}$, distance between camera and ground $d$, acceptable inlier threshold $\epsilon$
		\Ensure space convergence signal $c^{\star}$, lower bound value $v_{lb}$, upper bound value $v_{ub}$, 2D-2D correspondences

		\State Calculate the lower bound by evaluating the objective at the interval centre
		\State Compute the upper bound by deriving the transfer bounding box for each point in $\mathcal{P}_2$ and summing up the number of points from $\mathcal{P}_1$ that lie within the box
		\State \textbf{if} $v_{lb} == v_{ub}$ \textbf{then} $c^{\star} = 1$ \textbf{else} $c^{\star} = 0$
	\end{algorithmic}
	\label{Space_Algorithm}
\end{algorithm}

Starting from an initial domain, the algorithm consecutively divides 2-dimensional intervals into four subspaces, each time evaluating lower and upper bounds and pruning intervals if their upper bound is lower than the lower bound of another interval. We terminate the branching when the interval reaches its convergence signal where the lower bound equals the upper bound. The branch-and-bound framework can lead to a worst-case time and space complexity of $O(2^N)$, where $N$ represents number of iterations. In the following, we introduce further details on the search strategy, the initialisation, and the termination condition which help to speed up the execution.

\textit{Search strategy:} The obvious choices are a breadth-first or depth-first strategy, or a combination of both by defining a clever heuristic. The depth-first strategy is uncertain to improve computational efficiency as it may start to search in a wrong part of space, and thus carve out a local optimum that is actually not useful to quickly prune other parts of the optimisation space. We adopt the breadth-first search strategy to examine the space level-by-level, and apply parallelisation to speed up the execution.

\textit{Initialisation:} We initialise the search in a smartly-chosen, small interval which largely reduces computational load. In practice, the vehicle's movement is a smooth, continuous process. Given fixed integration times, $\phi$ and $\rho$ are directly linked to the vehicle's rotational and translational velocity, and as such undergo smooth variation for themselves. We therefore record the optimal solution from the previous pair of views, and initialize the search with a smaller interval centered around this optimum.

\textit{Termination:} There may be ambiguity upon convergence. Multiple branches on the same level could reach the same objective energy (i.e. the value corresponding to both upper and lower bound). However, these branches are likely to form a cluster and share very similar correspondences. We resolve the ambiguity and speed up the computation by actively identifying situations in which multiple neighbouring branches share almost same correspondences, and use the latter for non-linear refinement of $\phi$ and $\rho$.


\section{EXPERIMENTAL RESULTS}
\label{sec:experiments}

We perform two sets of experiments. We start with simulation experiments that test the algorithm's ability to handle noise under different conditions. We also test the influence of violations of the Ackermann model. The section concludes with experiments on real data, demonstrating how our method is able to outperform state-of-the-art correspondence-based solutions. All experiments are conducted on an Intel Core i7-8550U 1.8 GHz CPU with 16GB RAM.

\subsection{Synthetic Data}

We create our simulation experiments by defining a virtual, 4$\times$4$m^2$ canvas on which we uniformly sample a certain number of points. We then generate virtual feature measurements by assuming the existence of a perspective camera that travels along a horizontal trajectory with a perfectly downward directed principal axis. The distance between the camera and the canvas is set to 0.2$m$, and the camera has VGA resolution and a focal length of 500pix.

In our first experiment, the camera is set to move along simple circular trajectories, thus probing algorithm consistency. Random uniform noise between -2.5pix and 2.5pix is added to each coordinate of each virtual feature measurement. The total number of points on the entire canvas is also varied between 3000 and 9000 points. Our result is indicated in Figure \ref{fig:simulation}, where we depict the mean Relative Pose Error (RPE) as defined in \cite{sturm2012benchmark}. Figure \ref{fig:simulation:before} shows the error before non-linear refinement, where we simply take the average of the centers of all intervals that remained after convergence (intervals for which upper bound equals to lower bound). Figure \ref{fig:simulation:after} shows the result obtained by further refining this average based on all final correspondences. The refinement consists of a simple non-linear optimisation procedure in which we minimise the Euclidean transfer error. As can be observed, the algorithm has a good ability to overcome noise. We can furthermore conclude that good accuracy depends on sufficient features. Having about 20 inlier correspondences between successive frames proves to lead to a good result, while injecting more features will generally not improve the result any further.

In our second experiment, we analyse deviations from the Ackermann model. The circular trajectory is replaced by an ellipse for which $a = 1.0m$ and $b$ is varied between 0.6$m$ and 1.0$m$, thus leading to changing eccentricity $m = \sqrt{1 - b^2/a^2}$. Though we make sure that the camera $y$ axis stays tangential to the ellipse, this leads to displacements along non-circular arcs and thus varying violations of the Ackermann motion model. The validity however also depends on the baseline between the views. As indicated in Figure \ref{fig:simulation:eccentricity}, lower values for $b$ lead to higher eccentricities, which we effectively encounter by reducing the baseline. The baseline is changed as a function of a step size $x$ used to sample $360/x$ equally spaced frames along the ellipse. Note that in practice smaller baselines are achieved by higher framerates.

\subsection{Real Data}

To conclude, we also test our algorithm on real data. Images are captured by a camera (DALSA G3-GC11-C1920, with image size of 1936$\times$1216) mounted underneath an indoor AGV with an artificial support light. The framerate of this camera is set to 12fps, restricted by the frequency of the light. The wide angle lens (Myutron FV0420) has a focal length of 4$mm$. The distance between the camera and the ground is about 0.25$m$. The vehicle speed ranges between 0 and 0.2$m/s$. We tested a large collection of possible feature detectors, and converged onto the FAST feature detector \cite{rosten2008faster} as it produces sufficiently many and robust features at an acceptable computational expense (properties such as scale and rotation invariance are not a requirement in our application). We furthermore use a state-of-the-art feature tracker to establish correspondences between successive views, and apply the 1 Point Ransac algorithm presented in \cite{scaramuzza2009real} with the same non-linear refinement. This serves well as a more traditional reference implementation as the method also relies on the Ackermann model and is generally believed to exhibit very high robustness with respect to outlier correspondences. Absolute Trajectory Error (ATE) and Relative Pose Error (RPE) result as defined in \cite{sturm2012benchmark} are indicated in Table \ref{table_GOVO}. As can be observed, the 1-point Ransac algorithm is unable to reliably estimate the trajectory of the vehicle, thus pointing at the difficulty of feature matching. On the other hand, our correspondence-less algorithm is still able to accurately estimate the motion. Our algorithm runs between 5 and 15Hz on a regular CPU machine, and this therefore shows a capability of near real-time execution.

\begin{table}
\vspace{0.2cm}
\renewcommand{\arraystretch}{1.2}
\caption{performance comparison}
\label{table_GOVO}
\centering
\begin{tabular}{|c|c|c|c|c|}
\hline
& \multicolumn{2}{|c|}{GOVO} & \multicolumn{2}{|c|}{1 Point RANSAC} \\
\cline{2-5}
& RPE [m] & ATE [m] & RPE [m] & ATE [m] \\
\hline
Head Straight & \textbf{0.1425} & \textbf{0.0653} & 1.7622 & 0.3112 \\
\hline
Take Turns & \textbf{0.4847} & \textbf{0.1352} & 1.0717 & 0.4035 \\
\hline
Combination & \textbf{1.0032} & \textbf{0.5824} & 4.0688 & 1.6519 \\
\hline
\end{tabular}
\end{table}


\section{CONCLUSIONS}

We have presented the first globally optimal and correspondence-less solution to planar motion estimation in front of a planar scene, and successfully applied it to odometry estimation for planar ground vehicles. While this is interesting in itself, there is a deeper message behind our contribution. The difficulty of extracting reliable feature descriptors from ground plane textures motivates the use of the correspondence-less branch-and-bound optimisation technique. However, such methods typically suffer from the curse of dimensionality and are hardly amenable to real-time execution. We demonstrate that this is not necessarily a problem in the case of planar motion, which may be described by the low-dimensional Ackermann steering model. To the best of our knowledge, our work is the first to demonstrate a near real-time application of the branch-and-bound technique for image registration. Given the small depth-of-scene, it is necessary to process images at a very high frame rate if elevated vehicle speeds are to be estimated. Therefore, our future efforts consist of further improving computational efficiency and employing event-based cameras to obtain very low-latency perception, and explore its potential towards accurate visual odometry in dynamic scenarios in conjunction with our globally optimal correspondence-less registration approach.


\section*{Acknowledgments}

The authors would like to thank the fundings sponsored by Natural Science Foundation of Shanghai (grant number: 19ZR1434000) and Natural Science Foundation of China (grant number: 61950410612).


\addtolength{\textheight}{-5.9cm} 



\end{document}